# Integrating Innate and Adaptive Immunity for Intrusion Detection

Gianni Tedesco[1], Jamie Twycross[1], and Uwe Aickelin[1]

School of Computer Science & IT (ASAP)
University of Nottingham
NG8 1BB
gxt,jpt,uxa@cs.nott.ac.uk

**Abstract.** Network Intrusion Detection Systems (NIDS) monitor a network with the aim of discerning malicious from benign activity on that network. While a wide range of approaches have met varying levels of success, most IDS's rely on having access to a database of known attack signatures which are written by security experts. Nowadays, in order to solve problems with false positive alerts, correlation algorithms are used to add additional structure to sequences of IDS alerts. However, such techniques are of no help in discovering novel attacks or variations of known attacks, something the human immune system (HIS) is capable of doing in its own specialised domain. This paper presents a novel immune algorithm for application to an intrusion detection problem. The goal is to discover packets containing novel variations of attacks covered by an existing signature base.

## 1 Introduction

Network intrusion detection systems (NIDS) are usually based on a fairly low level model of network traffic. While this is good for performance it tends to produce results which make sense on a similarly low level which means that a fairly sophisticated knowledge of both networking technology and infiltration techniques is required to understand them.

Intrusion alert correlation systems attempt to solve this problem by post-processing the alert stream from one or many intrusion detection sensors (perhaps even heterogeneous ones). The aim is to augment the somewhat one-dimensional alert stream with additional structure. Such structural information clusters alerts in to "scenarios" - sequences of low level alerts corresponding to a single logical threat.

A common model for intrusion alert correlation algorithms is that of the attack graph. Attack graphs are directed acyclic graphs (DAGs) that represent the various types of alerts in terms of their prerequisites and consequences. Typically an attack graph is created by an expert from a priori information about attacks. The attack graph enables a correlation component to link a given alert with a previous alert by tracking back to find alerts whose consequences imply the current alerts prerequisites. Another feature is that if the correlation algorithm is run in reverse, predictions of future attacks can be obtained.

In implementing basic correlation algorithms using attack graphs, it was discovered that the output could be poor when the underlying IDS produced false negative alerts. This could cause scenarios to be split apart as evidence suggestive of a link between two scenarios is missing. This problem has been addressed in various systems [8, 6] by adding the ability to hypothesise the existence of the missing alerts in certain cases. [7] go as far as to use out of band data from a raw audit log of network traffic to help confirm or deny such hypotheses.

While the meaning of correlated alerts and predicted alerts is clear, hypothesised results are less easy to interpret. Presence of hypothesised alerts could mean more than just losing an alert, it could mean either of:

1. The IDS missed the alert due to some noise, packet loss, or other low level sensor problem
2. The IDS missed the alert because a novel variation of a known attack was used
3. The IDS missed the alert, because something not covered by the attack graph happened (totally new exploit, or new combination of known exploits)

This work is motivated specifically by the problem of finding novel variations of attacks. The basic approach is to apply AIS techniques to detect packets which contain such variations. A correlation algorithm is taken advantage of to provide additional safe/dangerous context signals to the AIS which would enable it to decide which packets to examine. The work aims to integrate a novel AIS component with existing intrusion detection and alert correlation systems in order to gain additional detection capability.

## 2 Background

### 2.1 Intrusion Alert Correlation

Although the exact implementation details of attack graphs algorithms vary, the basic correlation algorithm takes an alert and an output graph, and modifies the graph by addition of vertices and/or edges to produce an updated output graph reflecting the current state of the monitored network system.

For the purposes of discussion, an idealised form of correlation output is defined which hides specific details of the correlation algorithm from the AIS component. This model, while fairly simple, adequately maps to current state of the art correlation algorithms.

Firstly, as in [8], exploits are viewed as a 3-tuple $(vuln, src, dst)$ where vuln is the identity of a know exploit and src and dst refer to two hosts which must be connected for the exploit to be carried out accross the network. An injective function "f" $(ALERT \rightarrow EXPLOIT)$. This is because there may be several variations of a single exploit, each requiring a different signature from the underlying IDS and consequently producing distinct alerts. Parenthetically, many IDS signatures contain within them meta-data such as the Bugtraq or Mitre Common Vulnerabilities and Exposures (CVE) identification numbers which allows this function to be implemented automatically.

With our assumptions stated we may proceed to define our correlation graph. The output graph, G, is defined as a DAG with exploit vertices ($V_e$), condition vertices ($V_c$) and edges (E):

$G = V_e \cup V_c \cup E$

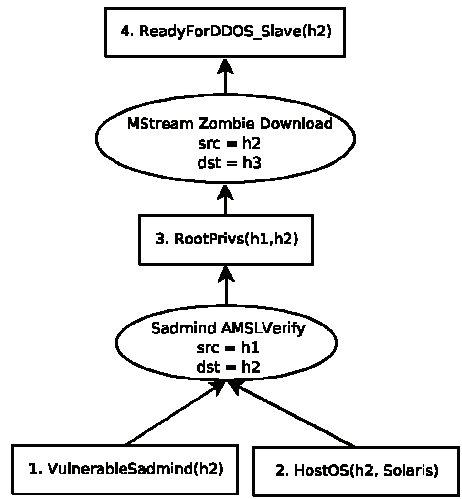

Fig. 1. Example output graph (conditions are boxes and exploits are ellipses).

The two types of vertex are necessary for being able to represent both conjunctive and disjunctive relations between exploits. As we can imagine by looking at Figure 1, any number of exploits may lead to condition 3, namely compromise of root privileges. This would mean that either the "AMSLVerify" exploit or some other root exploit may precede "Mstream Zombie Download." In another situation we may want "AMSLVerify" and some other exploit to be the precondition. In this case we would simply introduce another pre-requisite condition for that exploit alongside condition 3.

Each disconnected subgraph is considered as a threat scenario. That is to say, a structured set of low level alerts which constitute a single logical attack sequence.

There is a function "vertexstate" ($VERTEX \rightarrow VERTEX\ STATE$) which returns a 3 valued type, $\{HYP, REAL, PRED\}$ for hypothesised, real or pre-

dicted respectively. Condition vertices have a function "val" (V ERT EX →
BOOL) which tells us the value of the condition.

In addition to this, exploit vertices contain information about the computer
systems involved. Functions for retrieving source and destination addresses and
ports are also provided. For the purposes of discussion we will assume that
monitored networks are using the familiar TCP/IP protocol suite. Consequently
we shall refer to these functions as "src", "dst", "srcport" and "dstport."

## 2.2 Danger Theory

Over the last decade the focus of research in immunology has shifted from the
adaptive to innate immune system, and the cells of innate immunity has moved
to the fore in understanding the behaviour of immune system as a whole[2].
Insights gained from this research are beginning to be appreciated and modelled
at various levels by researchers building artificial immune systems.

The algorithm described in Section 3 incorporates at a conceptual level mechanisms from both the innate and adaptive immune system although, because of
the change in problem domain, these are implemented differently. This section
briefly reviews the biological processes and mechanisms which have been drawn
upon when designing the algorithm presented in this paper.

The biological immune system as a whole provides effective host defense
through the complex interaction of various immune system cells with themselves
and their environment, the tissue of the host organism. Dendritic cells (DCs),
part of the innate immune system, interact with antigen derived from the host
tissue and control the state of adaptive immune system cells.

Antigen is ingested from the extracellular milieu by DCs in their immature
state and then processed internally. During processing, antigen is segmented and
attached to major histocompatibility complex (MHC) molecules. This MHC-antigen complex is then presented under certain conditions on the surface of the
DC. As well as extracting antigen from their surroundings, DCs also have receptors which respond to a range of other signalling molecules in their milieu. Certain molecules, such a lipopolysaccaride, collectively termed pathogen-associated
molecular proteins (PAMPs[3]) are common to entire classes of pathogens and
bind with toll-like receptors (TLRs) on the surface of DCs.

Other groups of molecules, termed danger signals, such as heat shock proteins
(HSPs), are associated with damage to host tissue or unregulated, necrotic cell
death and bind with receptors on DCs. Other classes of molecules related to
inflammation and regulated, apoptotic cell death also interact with receptor
families present on the surface of DCs. The current maturation state of the DC
is determined through the combination of these complex signalling networks.
DCs themselves secrete cell-to-cell signalling molecules called cytokines which
control the state of other cell types. The number and strength of DC cytokine
output depends on its current maturation state.

T-cells, members of the adaptive immune system, have receptors which bind
to antigen presented in an MHC-antigen complex on the surface of DCs and respond to the strength of the match between receptor and antigen. This response is usually a change in the differentiation state of the T-cell. However, this response is not solely determined by antigen, but also by the levels of cytokines sensed by a range of cytokine receptors present on the surface of T-cells. These receptors are specific for cytokines produced by DCs.

In summary, DCs uptake and present antigen from the environment to T-cells. Also, DCs uptake signals from the environment and produce signals which are received by T-cells. The ultimate response of a T-cell to an antigen is determined by both the antigen presented by the DC and the signals produced by the DC. Section 3 below describes the implementation of this model in the context of a computer intrusion detection problem.

## 3 The Algorithm

For this purpose the "libtissue" [9, 10] AIS framework, a product of a danger theory project [1], will model a number of innate immune system components such as dendritic cells in order to direct an adaptive T-cell based response. Dendritic cells will carry the responsibility of discerning dangerous and safe contexts as well as carrying out their role of presenting antigen and signals to a population of T-cells as in [4].

**Tissue and Dendritic Cells** Dendritic cells (henceforth DCs) are of a class of cells in the immune system known as antigen presenting cells. They differ from other cells in this class in that this is their sole discernible function. As well as being able to absorb and present antigenic material DCs are also well adapted to detecting a set of endogenous and exogenous signals which arise in the tissue (IDS correlation graph).

These biological signals are abstracted in our system under the following designations:

1. Safe: Indicates a safe context for developing toleration.
2. Danger: Indicates a change in behaviour that could be considered pathological.
3. Pathogen Associated Molecular Pattern (PAMP)[3]: Known to be dangerous. In our system a distinction is made between activation by endogenous danger signals or through TLR receptors.

All of these environmental circumstances, or inputs, are factors in the life cycle of the DC. In the proposed system, DCs are seen as living among the IDS environment. This is achieved by wiring up their environmental inputs to changes in the IDS output state. A population of DCs are tied to the prediction vertices in the correlation graph, one DC for each predicted attack. Packets matching the prediction criteria of such a vertex are collected as antigen by the corresponding DC. These packets are either stored in memory or logged to disk until the DC matures and is required to present the antigen to a T-cell.

Once a prediction vertex has been added to the correlation graph, the arrival of subsequent alerts can cause that vertex to either be upgraded to an exploit vertex, changed to a hypothesised vertex, or become redundant as sibling vertices are so modified. These possible state changes will result in either a PAMP, danger or safe signal respectively.

These signals initiate maturation and consequent migration of the DC to a virtual lymph node where they are exposed to a population of T-cells.

The signal we are most interested in is the PAMP signal, this occurs when a predicted vertex becomes hypothesised. This provides us with a counterfactual hypothesis to test, ie. "suppose a novel a variation of the attack was carried out." The hypothesis is not unreasonable since:

1. The exploit was predicted already therefore it's prerequisites are met.
2. An exploit which depends on the consequences of the attack was carried out therefore the consequences of the exploit are met.

However this is not enough for a proof, since the standard caveats about the accuracy of the model hold. An attacker may, after all, attempt an attack whose preconditions are not met, the attack will fail, but the IDS cannot know.

**Antigen Representation** An important part of the design of an AIS is the representation of the domain data. A number of choices are available [12, 13]. For this algorithm we chose to use a natural encoding for the problem domain.

Network packets are blobs of binary data, each one is decoded by the IDS. The decoding process involves extracting, interpreting and validating the relevant features for the purpose of matching the packet against the signature database.

Our proposed algorithm represents each packet as an array of (feature,val) tuples. The array contains a tuple for all possible features and is ordered by feature. Features can be either integers or character strings. Values may be set to wildcards if the corresponding feature is not present in the packet.

This approach imposes a total order on the features. Such an order may be based, for example, on position in the packet which in nearly all cases is invariant and defined in protocol specifications.

Note that this representation shares structural similarities with the actual signatures used in network IDS's. The connection is elaborated in the following sub-section.

**T-cells** By the time a DC in our system has received a PAMP signal, matured, migrated to a lymph node and bound to a T-cell it contains a number of candidate packets (our antigen) and an indication of which signal caused migration. The simple T-cell model outlined in this paper only incorporates DC's activated by PAMPs.

The problem here is to select a subset of packets which may contain the novel variation(s) we are looking for. The inverse of the "f" function in our correlation algorithm provides a number of candidate signatures which may be used as a starting point. Thus the additional context is used to significantly reduce the search space in this phase of the algorithm.

In order to find these possible variations, a version of the IDS signature matching algorithm is required which provides meaningful partial matching. Since most signatures entail string searching or regular expression matching this is not a trivial task. For now, it will suffice to simply sum the number of matching criteria in each signature for each packet. If a match is sufficiently close, all the relevant data is output for further analysis. Since most signatures have less than 10 criteria, this may not be effective in all cases, due to the anticipated difficulty in selecting good matching thresholds.

## 4  Experimental Results

In order to test the algorithm it is important to know how greatly the set of candidate packets for novel attack variations can be reduced. We perform a simple experiment to validate the algorithm in this way. We chose to prototype the algorithm inside Firestorm[14], a signature matching IDS which uses the de-facto standard snort[15] signatures.

A circa 2000 wu-ftpd[11] exploit called "autowux" is to be our novel variation on the snort "FTP EXPLOIT format string" signature (figure 2). These exploits share the same attack methodology, namely exploiting format string overflows in the File Transfer Protocol (FTP) "SITE EXEC" command.

```
alert tcp $EXTERNAL_NET any -> $HOME_NET 21 (msg:''FTP EXPLOIT format
string''; flow:to_server,established; content: ''SITE EXEC |25 30 32 30
64 7C 25 2E 66 25 2E 66 7C 0A|''; depth: 32; nocase;)
```

Fig. 2. Generic snort signature for FTP format string exploits

The IDS is loaded with a full signature set and is tested to make sure that the autowux exploit packets are not already detected. A contrived attack graph with 3 exploits is also created (see figure 3). An nmap scan is the prerequisite and vulnerability to rootkit installation is the consequence of our "novel" FTP exploit.

The attack scenario is successfully played out across an otherwise quiet test network (run #1). The attack contains on the order of three thousand packets and the problem should be fairly simple because in the absence of background noise a high proportion of the packets are part of the FTP attack (975 of them to be precise). To make things more realistic, a second run of the experiment is carried out in which there is background FTP traffic to our vulnerable host. The background traffic is from the Lincoln Labs FTP data-set[16].

The two data sets were merged based on time deltas between packets, the start packets are synchronised. This provides a realistic and repeatable mix of benign and attack traffic (run #2).

The table below gives initial results for the prototype implementation based on a number of uncontrolled experiments. Total packets is the total number of packets in the merged data set, Ag packets refers to candidate packets in the DC and output packets refers to the final results - ie. those packets in which there is a suspected novel variation of an attack. False positive (FP) and false negative (FN) rates are calculated through manual analysis of the output. In this case, there is one true positive in each data set so all candidate output packets that are not true positives are false positives, so the rate is calculated with $\frac{n-1}{n}$.

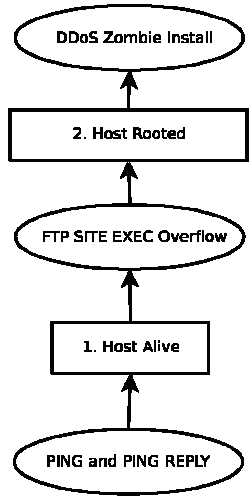

Fig. 3. Contrived attack graph used for experimental purposes

Run #1 is performed on a quiescent network, run #2 is with background traffic as described above.

| Run | Total Packets | Ag Packets | Output Packets | FP Rate | FN Rate |
|---|---|---|---|---|---|
| #1 | 3,000 | 975 | 18 | 94% | 0% |
| #2 | 18,000 | 8,000 | 30 | 96% | 0% |

Table 1. Accuracy of Algorithm with and without Background Traffic

The table shows that the packets of interest are extracted (eg. 975 / 3000) and that after further processing this is reduced to a mere handful of packets (eg. 18). Overall the detection rate is good, despite the high false positive rates (eg. 94%) which are inherent in the problem.

## 5   Conclusions and Future Work

In summation, a novel intrusion detection algorithm was presented drawing on theoretical models of innate immunity. The algorithm incorporates within it existing IDS algorithms, but expands on their capability in a limited area, detection of unknown (or 0-day) attacks which are based on other attacks that are previously known to the IDS. The AIS neatly interfaces with the problem domain by treating internal IDS data structures as an artificial tissue environment. Finally the algorithm was evaluated in terms of how accurately the novel variations can be identified.

It should be noted that the results are not directly comparable to other IDS algorithms as the problem being solved is uniquely circumscribed. Rather than designing an anomaly detection algorithm to find previously unknown attacks, a misuse detector and alert correlator are extended to detect a certain kind of anomaly arising from the incomplete models that are invariably used with such algorithms.

Initial results are promising despite the high false positive rate. However since the output is already clustered (all packets which were in a given DC are linked together) it means that as long as there is an upper bound on false positives and the false negative rate is low, there will usually be an accurate detection among each such cluster.

The DCs in the presented model are able to detect specific anomalous patterns of tissue growth and identify where and when novel attacks are taking place. After a DC has made an initial selection of candidate packets, it is then the responsibility of the T-cells to reduce the number of packets still further by detecting structural similarities in the data. DCs are concerned primarily with detecting abnormal behaviour within their environmental context, whereas T-cells are concerned primarily with discerning patterns within the antigen data. The co-ordination of both types of immune cell with each other and the tissue through orthogonal programming interfaces make for neat and efficient solution.

Further investigation in to the T-cell phase of the algorithm should be fruitful. The algorithm presented in this paper is fairly basic and does not incorporate meaningful partial matching which is important for performance and accuracy. A tolerance mechanism might also be useful in integrating the information conveyed by the safe and danger signals to further improve the false positive rate in the difficult cases where malicious traffic differs only slightly from legitimate traffic. Future testing should also incorporate historically problematic attack variations in order to provide a more realistic appraisal of the algorithm.

A mechanism for the automated generation of signatures for the novel variations discovered by the algorithm would be ideal. Work such as [17] shows us that this should, in theory, be possible with acceptable precision.

2003.
2. R N Germain. "An innately interesting decade of research in immunology". Nature Medicine. Vol. 10, No. 4, pp. 1307-1320. 2004.
3. CA Janeway Jr. "Approaching the Asymptote? Evolution and Revolution in Immunology." Cold Spring Harb Symp Quant Biol. 54 Pt 1:1-13. 1989.
4. J Greensmith, U Aickelin and S Cayzer. "Introducing Dendritic Cells as a Novel Immune-Inspired Algorithm for Anomaly Detection." 4th International Conference on Articial Immune Systems, pp 153-167. 2005.
5. P Matzinger. "Tolerance, danger and the extended family." Annual Reviews in Immunology, 12:991-1045, 1994.
6. P Ning. D Xu. "Hypothesizing and Reasoning about Attacks Missed by Intrusion Detection Systems." ACM Transactions on Information and System Security Vol. 7, No. 4, pp. 591-627. 2004.
7. P Ning, D Xu. CG Healey and R St. Amant. "Building Attack Scenarios through Integration of Complementary Alert Methods" Proceedings of the 11th Annual Network and Distributed System Security Symposium. 2004.
8. L Wang, A Liu and S Jajoda. "An Efficient Unified Approach to Correlating Hypothesising, and Predicting Intrusion Alerts." Proceedings of European Symposium on Computer Security. 2005.
9. J Twycross and U Aickelin. "Towards a Conceptual Framework for Innate Immunity." Proceedings of the 4th International Conference on Artificial Immune Systems, Banff, Canada, 2005.
10. J Twycross and U Aickelin. "libtissue - implementing innate immunity." Proceedings of the Congress on Evolution Computation. 2006.
11. Washington University FTP Server. http://www.wu-ftpd.org/
12. K. Mathias and D. Whitley. "Transforming the Search Space with Gray Coding." IEEE Conf. on Evolutionary Computation. Volume 1. pp: 513-518, 1994.
13. J Balthrop, F Esponda, S Forrest, M Glickman. "Coverage and Generalization in an Artificial Immune System." Genetic and Evolutionary Computation Conference (GECCO) 2002.
14. G Tedesco. Firestorm Network Intrusion Detection System. http://www.scaramanga.co.uk/firestorm/.
15. M Roesch. Snort Network Intrusion Detection System. http://www.snort.org/
16. Berkeley Labs Internet Traffic Archive Data Set: LBNL-FTP-PKT. http://www-nrg.ee.lbl.gov/LBNL-FTP-PKT.html
17. V Yegneswaran, JT Giffin, P Barford, S Jha. "An Architecture for Generating Semantics-Aware Signatures." Proceedings of USENIX Security Conference, 2005.